\documentclass[10pt,twocolumn,letterpaper]{article}

\usepackage{cvpr}
\usepackage{times}
\usepackage{epsfig}
\usepackage{graphicx}
\usepackage{amsmath}
\usepackage{amssymb}
\usepackage{booktabs}
\usepackage{placeins}

\usepackage{hyperref}
\hypersetup{
    colorlinks=true,
    linkcolor=blue,
    filecolor=magenta,      
    urlcolor=magenta,
}
 
\cvprfinalcopy 

\def\cvprPaperID{5755} 

\begin{document}

\title{Fusing Wearable IMUs with Multi-View Images for Human Pose Estimation: A Geometric Approach}
\author{Zhe Zhang\textsuperscript{1}\thanks{Work done when Zhe Zhang is an intern at Microsoft Research Asia.}\\
\and
Chunyu Wang\textsuperscript{2}\\
\and
Wenhu Qin\textsuperscript{1}\\
\and
Wenjun Zeng\textsuperscript{2}\\
\and
\textsuperscript{1}Southeast University, Nanjing, China
\and
\textsuperscript{2}Microsoft Research Asia, Beijing, China
}

\maketitle

\begin{abstract}
We propose to estimate $3$D human pose from multi-view images and a few IMUs attached at person's limbs. It operates by firstly detecting $2$D poses from the two signals, and then lifting them to the $3$D space. We present a geometric approach to reinforce the visual features of each pair of joints based on the IMUs. This notably improves $2$D pose estimation accuracy especially when one joint is occluded. We call this approach Orientation Regularized Network (\emph{ORN}).  Then we lift the multi-view $2$D poses to the $3$D space by an Orientation Regularized Pictorial Structure Model (\emph{ORPSM}) which jointly minimizes the projection error between the $3$D and $2$D poses, along with the discrepancy between the $3$D pose and IMU {orientations}. The simple two-step approach reduces the error of the state-of-the-art by a large margin on a public dataset. Our code will be released at \url{https://github.com/CHUNYUWANG/imu-human-pose-pytorch}.

\end{abstract}

\section{Introduction}

Estimating $3$D poses from images has been a longstanding goal in computer vision. With the development of deep learning models, the recent approaches \cite{gall2010optimization,PavlakosZDD17,belagiannis20143D,Rhodin_2018_ECCV,rhodin2018learning,trumble2018deep} have achieved promising results on the public datasets. One limitation of the vision-based methods is that they cannot robustly solve the occlusion problem. 

A number of works are devoted to estimating poses from wearable sensors such as IMUs \cite{trumble2017total,roetenberg2009xsens,von2016human,von2017sparse}. They suffer less from occlusion since IMUs can provide direct $3$D measurements. For example, Roetenberg \etal \cite{roetenberg2009xsens} place $17$ IMUs with $3$D accelerometers, gyroscopes and magnetometers at the rigid bones. If the measurements are accurate, the $3$D pose is fully determined. In practice, however, the accuracy is limited by a number of factors such as calibration errors and the drifting problem.

Recently, fusing images and IMUs to achieve more robust pose estimation has attracted much attention \cite{trumble2017total,von2018recovering,gilbert2019fusing,malleson2017real}. They mainly follow a similar framework of building a parametric $3$D human model and optimizing its parameters to minimize its discrepancy with the images and IMUs. The accuracy of these approaches is limited mainly due to the hard optimization problem. 

\begin{figure}
	\centering
	\includegraphics[width=1\linewidth]{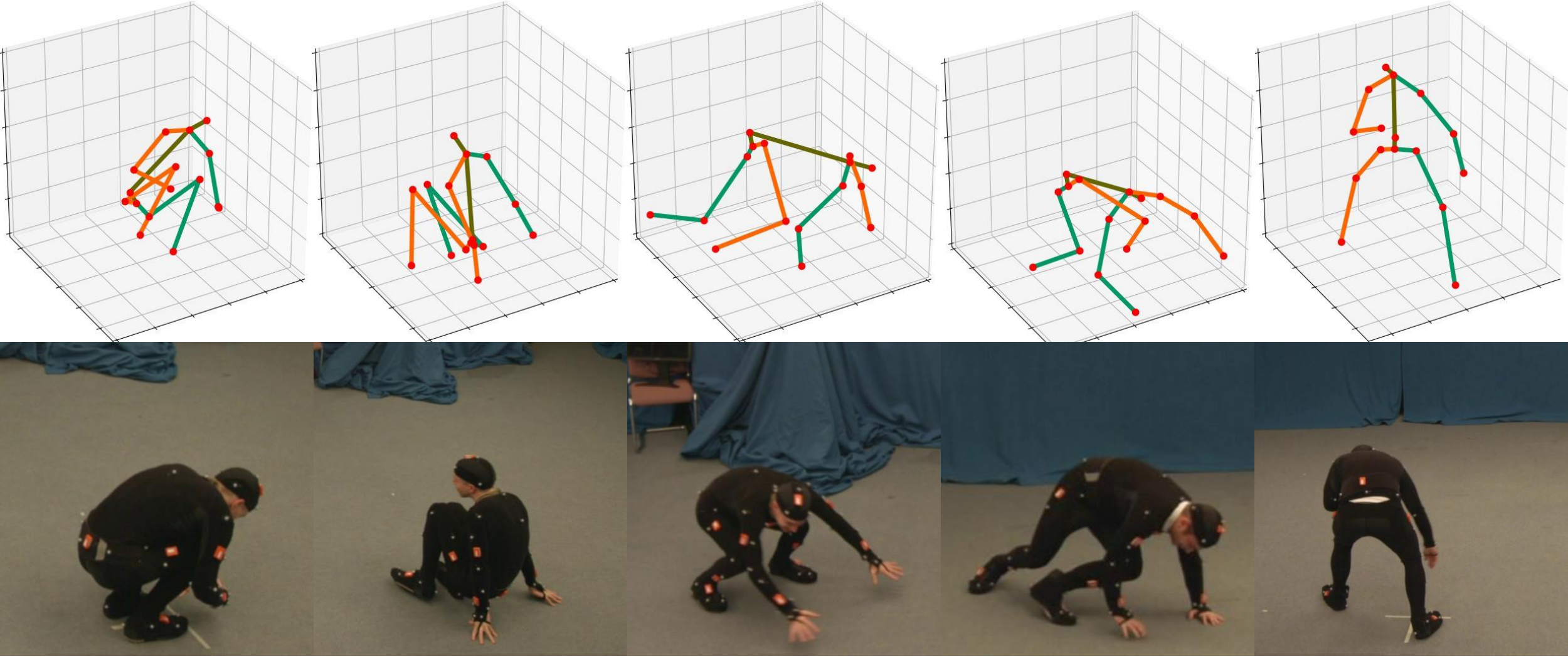}
	\caption{Our approach gets accurate $3$D pose estimations even when severe \emph{self-occlusion} occurs in the images.}
	\label{fig:sample}
\end{figure}

\begin{figure}
	\centering
	\includegraphics[width=0.95\linewidth]{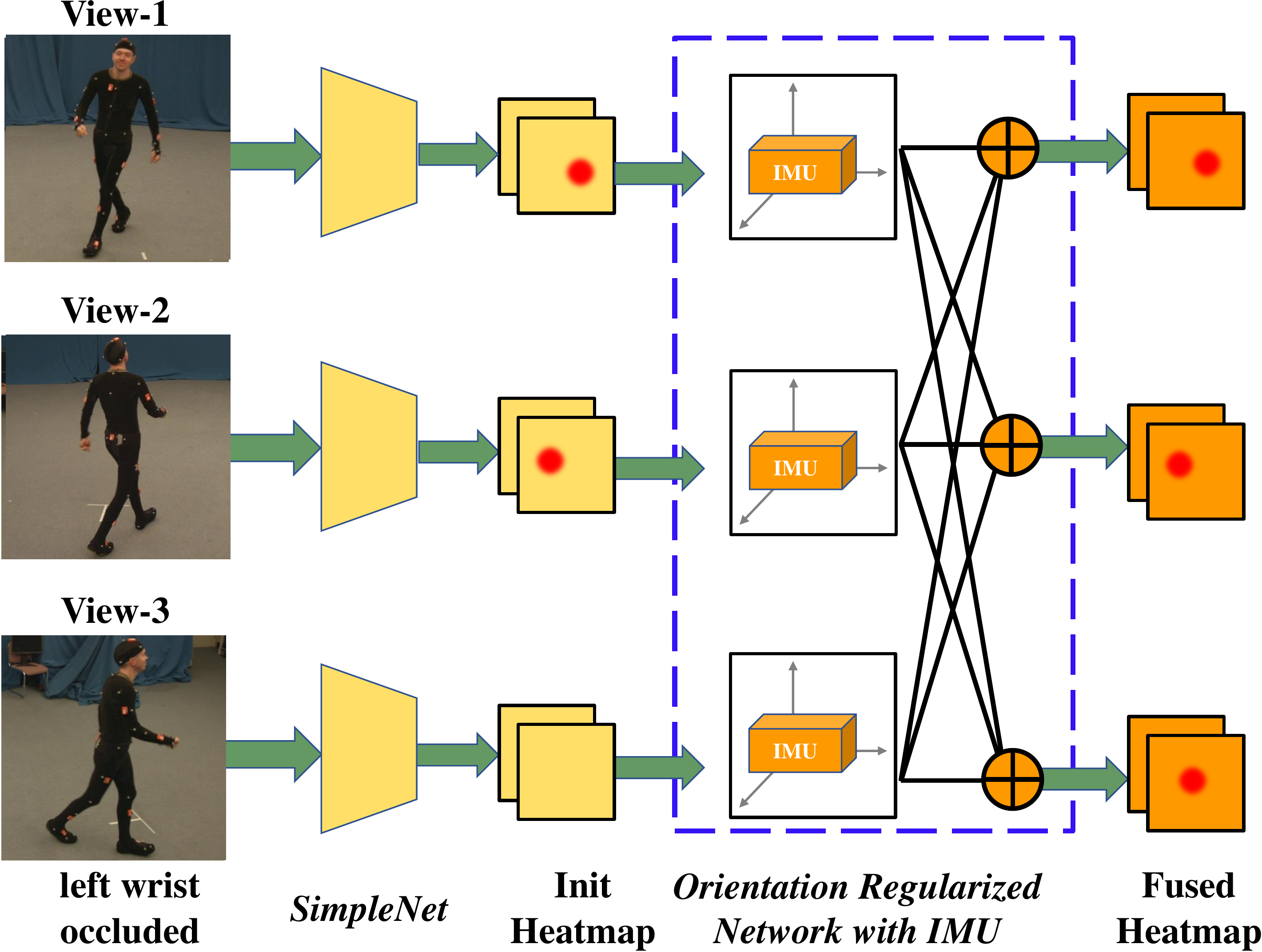}
	\caption{Overview of ORN. It firstly takes multi-view images as input and estimates initial heatmaps (based on SimpleNet \cite{simplebaselines}) independently for each camera view. Then with the aid of IMU orientations, it mutually fuses the heatmaps of the linked joints across all views. It enforces supervision on both initially estimated and fused heatmaps during the end-to-end training.}
	\label{fig:pipeline}
\end{figure}

We present an approach to fuse IMUs with images for robust pose estimation. It gets accurate estimations even when occlusion occurs (see Figure \ref{fig:sample}). In addition, it outperforms the previous methods \cite{malleson2017real, von2018recovering} by a notable margin on the public dataset. 
We first introduce Orientation Regularized Network (ORN) to \emph{jointly} estimate $2$D poses for multi-view images as shown in Figure \ref{fig:pipeline}. ORN differs from the previous multiview methods \cite{qiu2019cross} in that it uses IMU orientations as a structural prior to \emph{mutually} fuse the image features of each pair of joints linked by IMUs. For example, it uses the features of the elbow to reinforce those of the wrist based on the IMU at the lower-arm. 

The cross-joint-fusion allows to accurately localize the occluded joints based on their neighbors. The main challenge is to determine the relative positions between each pair of joints in the images, which we solve elegantly in the $3$D space with the help of IMU orientations. The approach significantly improves the $2$D pose estimation accuracy especially when occlusion occurs.

In the second step, we estimate $3$D pose from multi-view $2$D poses (heatmaps) by a Pictorial Structure Model (PSM) \cite{kostrikov2014depth,PavlakosZDD17,belagiannis20143D}. It jointly minimizes the projection error between the $3$D and $2$D poses, along with the discrepancy between the $3$D pose and the prior. The previous works such as \cite{PavlakosZDD17,qiu2019cross} often use the limb length prior to prevent from generating  abnormal $3$D poses. This prior is fixed for the same person and does not change over time. In contrast, we introduce an orientation prior that requires the limb orientations of the $3$D pose to be consistent with the IMUs. The prior is complementary to the limb length and can reduce the negative impact caused by inaccurate $2$D poses. We call this approach Orientation Regularized Pictorial Structure Model (ORPSM).

We evaluate our approach on two public datasets including Total Capture \cite{trumble2017total} and H36M \cite{ionescu2014human3}. On both datasets, \emph{ORN} notably improves the $2$D estimation accuracy especially for the frequently occluded joints such as ankle and wrist, which in turn decreases the $3$D pose error. Take the Total Capture dataset as an example, on top of the $2$D poses estimated by \emph{ORN}, \emph{ORPSM} obtains a $3$D position error of $24.6$mm which is much smaller than the previous state-of-the-art \cite{qiu2019cross} ($29$mm) on this dataset. This result demonstrates the effectiveness of our visual-inertial fusion strategy. To validate the general applicability of our approach, we also experiment on the H36M dataset which has different poses from the Total Capture dataset. Since it does not provide IMUs, we synthesize virtual limb orientations and only show proof-of-concept results.

\section{Related Work}
\paragraph{Images-based}

We classify the existing image-based $3$D pose estimation methods into three classes. The first class is model/optimization based \cite{gall2010optimization,liu2011markerless} which defines a $3$D parametric human body model and optimizes its parameters to minimize the discrepancy between model projections and extracted image features. These approaches mainly differ in terms of the used image features and optimization algorithms. These methods generally suffer from the difficult non-convex optimization which limits the $3$D estimation accuracy to a large extent in practice.

With the development of deep learning, some approaches such as \cite{Rhodin_2018_ECCV,rhodin2018learning,martinez2017simple,trumble2018deep,jafarian2018monet,pavllo20193d} propose to learn a mapping from images to $3$D pose in a supervised way. The lack of abundant ground truth $3$D poses is their biggest challenge for achieving desired performance on wild images. Zhou \etal \cite{zhou2017towards} propose a multi-task solution to leverage the abundant $2$D pose datasets for training. Yang \etal \cite{yang20183d} use adversarial training to improve the robustness of the learned model. Another limitation of this type of methods is that the predicted $3$D poses by these methods are relative to their pelvis joints. So they are not aware of their absolute locations in the world coordinate system.

The third class of methods such as \cite{amin2013multi,burenius20133D,PavlakosZDD17,belagiannis20143D,gilbert2018volumetric,joo2019panoptic,dong2019fast,qiu2019cross} adopt a two-step framework. It first estimates $2$D poses in each camera view and then recovers the $3$D pose in a world coordinate system with the help of camera parameters. For example, Tome \etal \cite{tome2018rethinking} build a $3$D pictorial model and optimize the $3$D locations of the joints such that their projections match the detected $2$D pose heatmaps and meanwhile the spatial configuration of the $3$D joints matches the prior pose structure. Qiu \etal \cite{qiu2019cross} propose to first estimate $2$D poses for every camera view, and then estimate the $3$D pose by triangulation or by pictorial structure model. This type of approaches has achieved the state-of-the-art accuracy due to the significantly improved $2$D pose estimation accuracy.

\paragraph{IMUs-based} There are a small number of works which attempt to recover $3$D poses using only IMUs. For example, Slyper \etal \cite{slyper2008action} and  Tautges \etal \cite{tautges2011motion} propose to reconstruct human pose from $5$ accelerometers by retrieving pre-recorded poses with similar accelerations from a database. They get good results when the test sequences are present in the training dataset. Roetenberg \etal \cite{roetenberg2009xsens} use $17$ IMUs equipped with $3$D accelerometers, gyroscopes and magnetometers and all the measurements are fused using a Kalman Filter. By achieving stable orientation measurements, the $17$ IMUs can fully define the pose of the subject. Marcard \etal \cite{von2017sparse} propose to exploit a statistical body model and jointly optimize the poses over multiple frames to fit orientation and acceleration data. One disadvantage of the IMUs-only methods is that they suffer from drifting over time, and need a large amount of careful engineering work in order to make it work robustly in practice.

\paragraph{``Images+IMUs''-based}
Some works such as \cite{von2016human,trumble2017total,von2018recovering,gilbert2019fusing,malleson2017real} propose to combine images and IMUs for robust $3$D human pose estimation. The methods can be categorized into two classes according to how image-inertial fusion is performed. The first class \cite{malleson2017real,von2018recovering,von2016human} estimate $3$D human pose by minimizing an energy function which is related to both IMUs and image features. The second class \cite{trumble2017total,gilbert2019fusing} estimate $3$D poses separately from the images and IMUs, and then combine them to get the final estimation. For example, Trumble \etal \cite{trumble2017total,gilbert2019fusing} propose a two stream network to concatenate the pose embeddings separately derived from images and IMUs for regressing the final pose. 

Although the simple two-step framework has achieved the state-of-the-art performance in the image only setting, it is \emph{barely} studied for ``IMU+images''-based pose estimation because it is nontrivial to leverage IMUs in the two steps. Our main contribution lies in proposing two novel ways of exploiting IMUs in the framework. More importantly, we empirically show that this simple two-step approach can significantly outperform the previous state-of-the-arts. 

Our work differs from the previous works \cite{trumble2017total,von2018recovering,gilbert2019fusing,malleson2017real,loper2015smpl} in two-fold. First, instead of estimating $3$D poses or pose embeddings from images and IMUs separately and then fusing them in a late stage, we propose to fuse IMUs and image features in a very early stage with the aid of $3$D geometry. This directly gives improved $2$D poses rather than attempting to get accurate poses from two inaccurate ones as in late fusion. Second, in the $3$D pose estimation step, we leverage IMUs in the pictorial structure model. Although pictorial model is not new, the effect of using IMUs has not been discussed. Finally, we hope this simple yet effective approach could promote more research in the two-step pose estimation direction.

\section{ORN for $2$D Pose Estimation}
We represent a $2$D pose by a graph which consists of $M$ joints $\mathcal{J}=\{{J}_1, {J}_2, \cdots, {J}_M \}$ and $N$ edges $\mathcal{E}=\{{e}_1, {e}_2, \cdots, {e}_N\}$ as shown in Figure \ref{fig:geometry} (c). Each ${J}$ represents the state of a joint such as its $2$D location in the image. Each edge ${e}$ connects two joints, representing their conditional dependence. In this work, we attach IMUs to ${W}$ limbs to obtain their $3$D orientations $\mathcal{O}=\{{o}_1, {o}_2, \cdots, {o}_W\}$. This orientation information will be used to constrain relative positions between two joints. In the following, we will describe in detail how we estimate $2$D poses with the help of orientations.

\subsection{Methodology}
We start by describing how $3$D limb orientations can be used to mutually enhance the features between pairs of joints linked by IMUs in the \emph{same} camera view. Then we extend it to handle multi-view features.
\begin{figure*}[tp]
	\centering
	\includegraphics[width=0.8\linewidth]{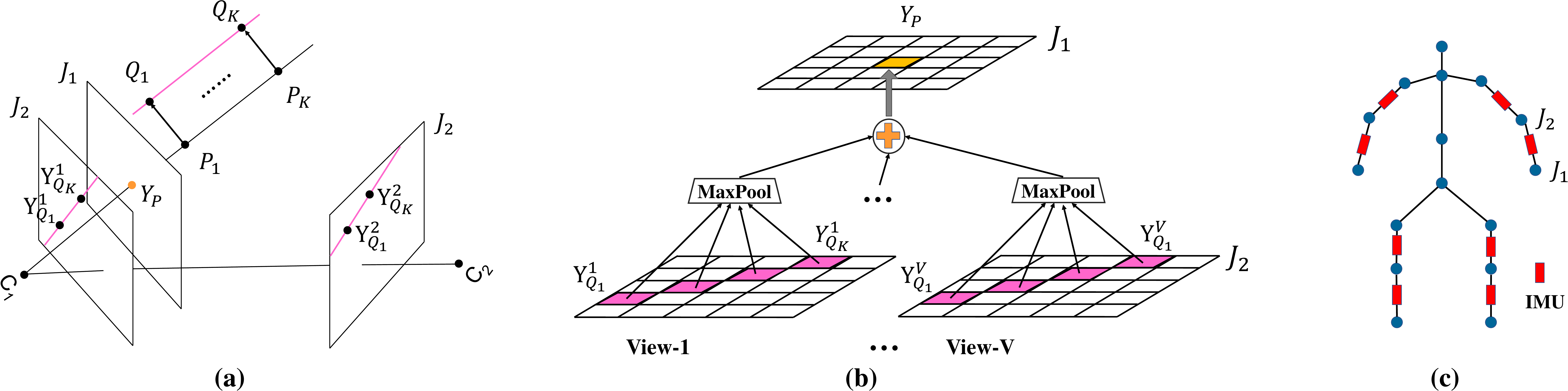}
	\caption{Illustration of the cross-joint-fusion idea in ORN. (a) For a location ${Y}_{P}$ in ${H}_1$, we estimate its $3$D points ${P}_k$ lying on the line defined by the camera center ${C}_1$ and ${Y}_{P}$. Then based on the $3$D limb orientation provided by IMU and the limb length, we get candidate $3$D locations of ${J}_2$ which are denoted as ${Q}_k$. We project ${Q}_k$ to the image as ${Y}_{{Q}_{k}}$ and get the corresponding heatmap confidence. If the confidence is high, ${J}_1$ has high confidence being located at ${Y}_{P}$. (b) We enhance the initial confidence of ${J}_1$ at ${Y}_{{P}}$ with the confidence of ${J}_2$ at ${Y}_{{Q}_k}$ in all views. Similarly, we can fuse the heatmap of ${J}_2$ using that of ${J}_1$. (c) We show the skeleton model used in this work. }
	\label{fig:geometry}
\end{figure*}

\paragraph{Same-View Fusion}
We explain the main idea of our approach with a pair of joints ${J}_1$ and ${J}_2$ as an example. The two joints are connected by the limb ${e}$ whose $3$D orientation is ${o}$. In practice, we will apply the fusion operation to all pairs of joints linked by IMUs. Figure \ref{fig:geometry} sketches the idea. Let the heatmaps of ${J}_1$ and ${J}_2$ be ${H}_1$ and ${H}_2$, respectively. For a location ${Y}_{{P}}$ in ${H}_1$, its heatmap value represents the confidence that ${J}_1$ is at ${Y}_{{P}}$. We propose to enhance it by the confidence of the linked joint ${J}_2$ at $K$ possible \emph{corresponding} locations ${Y}_{{Q}_k}, k=1, \cdots, K$ which are consistent with ${Y}_{{P}}$ according to limb orientation ${o}$.

The main challenge is to determine the locations of ${Y}_{{Q}_k}$. From Figure \ref{fig:geometry} (a), it is clear that the corresponding $3$D point ${P}$ of ${Y}_{{P}}$ has to lie on the line defined by the camera center ${C}_1$ and ${Y}_{{P}}$. Since the exact depth of ${P}$ is unknown, we log-uniformly sample $K$ locations  ${P}_k, k=1,\cdots,K$ on the line as its candidates \footnote{We use log-uniform instead of uniform sampling to prevent from generating redundant collapsed $2$D projections}. In addition, we assume the limb length $l$ between ${J}_1$ and ${J}_2$ is provided as a prior which is the average limb length computed on the training dataset. Together with the $3$D orientation ${o}$ between the two joints, we can compute the $3$D locations of ${J}_2$ as follows:
\begin{equation}
 {Q}_k={P}_k + {o} * l \quad \forall k=1,\cdots,K   
\end{equation}
Finally we project ${Q}_k$ onto the image using the camera parameters and get the $2$D locations as ${Y}_{{Q}_k}$. Intuitively, a high response at ${Y}_{{Q}_k}$ in ${H}_2$ actually indicates ${J}_1$ has a high probability to be at ${Y}_{{P}}$. This observation is the core of our fusion approach. However, there is ambiguity because we do not know which of the $K$ candidates ${Y}_{{Q}_k}$ is the corresponding point due to the lack of depth.

Our solution is to find the maximum response among all locations ${Y}_{{Q}_k},  k=1,\cdots,K$:
\begin{equation}
    {H}_1({Y}_{{P}}) \leftarrow \lambda {H}_1({Y}_{{P}}) + (1-\lambda) \max_{k=1 \cdots K}{{H}_2({Y}_{{Q}_k})}
\end{equation}
Since fusion happens in the heatmap layer, ideally, ${Y}_{{Q}_k}$ should have the largest response at the correct ${J}_2$ location and zeros at other locations. It means the non-corresponding locations will contribute no or little to the fusion. We set the balancing parameter $\lambda$ to be $0.5$ in our experiments. We sample $200$ points whose depths range from zero to the maximum depth value, which is determined by the size of the room.

\paragraph{Cross-View Fusion}
One limitation of the \emph{Same-View Fusion} is that the correct location ${Y}_{{Q}_{k^*}}$ which has the maximum response among the $K$ candidates in $H_2$, will contribute to multiple candidates like ${Y}_{{P}}$ in ${H}_1$. These candidates also lie on a line. But most of such locations do not correspond to the joint type ${J}_1$. In other words, some non-corresponding locations are mistakenly enhanced. For example, there are blurred lines in the ``Enhanced Heatmap'' in Figure \ref{fig:visualization_hm} with each from a different camera view.

To resolve this problem, we propose to perform fusion across multiple views simultaneously:
\begin{equation}
    {H}_1({Y}_{{P}}) \leftarrow \lambda {H}_1({Y}_{{P}}) +\frac{(1-\lambda)}{V} \sum_{v=1}^{V} \max_{k=1 \cdots K}{{H}_2^v{({Y}^v_{{Q}_k})}},
\end{equation}
where ${Y}^v_{{Q}_k}$ is the projection of ${Q}_k$ in the camera view $v$ and ${H}_2^v$ is the heatmap of ${J}_2$ in view $v$. The result is that the lines from multiple views will intersect at the correct location. Consequently, the correct location will be enhanced most which resolves the ambiguity. See the fused heatmap in Figure \ref{fig:visualization_hm} for illustration. Another desirable effect of cross-view fusion is that it helps solve the occlusion problem by fusing the features from multiple views because a joint occluded in one view may be visible in other views. This notably increases the joint detection rates.

\begin{table*}[]
\center
\caption{The $2$D pose estimation accuracy (PCKh@t) on the Total Capture Dataset. ``SN'' means SimpleNet which is the baseline. 
$\emph{ORN}^{same}$ and \emph{ORN}, respectively, represent that the same-view and cross-view fusion are used.  ``Mean (six)'' is the average result over the six joint types. ``Others'' is the average result over the rest of the joints. ``Mean (All)'' is the result over all joints.}
\label{table:detectionrate}
\begin{tabular}{lc||ccccccc||c||c}
\toprule
      Methods & PCKh@ & Hip    & Knee   & Ankle  & Shoulder & Elbow  & Wrist & \emph{Mean (Six)} & {Others} & Mean (All) \\ \hline
\emph{SN}    & 1/2  & 99.3 & 98.3 & 98.5 & 98.4 & 96.2 & 95.3 & 97.7 & 99.5 & 98.1 \\
$\emph{ORN}^{same}$  & 1/2  & 99.4 & 99.0 & 98.8 & 98.5 & 97.7 & 96.7 & 98.3 & 99.5 & 98.6\\
\emph{ORN} & 1/2  & \textbf{99.6} & \textbf{99.2} & \textbf{99.0} & \textbf{98.9} & \textbf{98.0} & \textbf{97.4} & \textbf{98.7} & 99.5 & {98.9}\\ \hline
\emph{SN}    & 1/6  & 97.5 & 92.3 & 92.5 & 78.3 & 80.8 & 80.0 & 86.9 & 95.4 & 89.1\\
$\emph{ORN}^{same}$  & 1/6  & 97.2 & 94.0 & 93.3 & 78.1 & 83.5 & 82.0 & 88.0 & 95.4 & 89.9\\
\emph{ORN} & 1/6  & \textbf{97.7} & \textbf{94.8} & \textbf{94.2} & \textbf{81.1} & \textbf{84.7} & \textbf{83.6} & \textbf{89.3} & 95.4 & {90.9}\\ \hline
\emph{SN}    & 1/12 & \textbf{87.6} & 67.0 & 68.6 & 47.4 & 50.0 & 49.3 & 61.7 & 78.1 & 65.8\\
$\emph{ORN}^{same}$  & 1/12 & 81.2 & 70.1 & 68.0 & 43.9 & 51.6 & 50.1 & 60.8 & 78.1 & 65.2\\
\emph{ORN} & 1/12 & 85.3 & \textbf{71.6} & \textbf{70.6} & \textbf{47.7} & \textbf{53.2} & \textbf{51.9} & \textbf{63.4} & 78.1 & {67.1}\\
\toprule
\end{tabular}
\end{table*}

\subsection{Implementation}
We use the network proposed in \cite{simplebaselines}, referred to as SimpleNet (SN) to estimate initial pose heatmaps. It uses ResNet50 \cite{he2016deep} as its backbone which was pre-trained on the ImageNet classification dataset. The image size is $256 \times 256$ and the heatmap size is $64 \times 64$. 
The orientation regularization module can either be trained end-to-end with SN, or added to a already trained SN as a plug-in since it has no learnable parameters. In this work, we train the whole ORN end-to-end.
We generate ground-truth pose heatmaps as the regression targets and enforce $l_\text{2}$ loss on all views before and after feature fusion. In particular, we do not compute losses for background pixels of the fused heatmap since the background pixels may have been enhanced.
The network is trained for $15$ epochs. The parameter $\lambda$ is $0.5$ in all experiments. Other hyper-parameters such as learning rate and decay strategy are the same as in \cite{simplebaselines}.

\section{ORPSM for $3$D Pose Estimation}
A human is represented by a number of joints $\mathcal{J}=\{{J}_1,
{J}_2,\cdots,{J}_M\}$. Each ${J}$ represents its $3$D position in a world coordinate system. Following the previous works \cite{kostrikov2014depth,PavlakosZDD17,belagiannis20143D,qiu2019cross}, we use the pictorial model to estimate $3$D pose as it is more robust to inaccurate $2$D poses. But different from the previous works, we also introduce and evaluate a novel limb orientation prior based on IMUs as will be described in detail later. Each ${J}$ takes values from a discrete state space. An edge between two joints denotes their conditional dependence such as limb length. Given a $3$D pose $\mathcal{J}$ and multi-view $2$D pose heatmaps $\mathcal{F}$, we compute the posterior as follows
\begin{equation}
\begin{split}
    p(\mathcal{J} | \mathcal{F})= &\frac{1}{Z(\mathcal{F})} \prod_{i=1}^M{\phi_i^{\text{conf}}({J}_i, \mathcal{F})}\prod_{(m, n) \in \mathcal{E}_{limb}}{\psi^{\text{limb}}({J}_m, {J}_n)} \\
    &\prod_{(m, n) \in \mathcal{E}_{IMU}}{\psi^{\text{IMU}}({J}_m, {J}_n)},
    \label{eq:psm}
\end{split}
\end{equation}
where $Z(\mathcal{F})$ is the partition function, $\mathcal{E}_{limb}$ and $\mathcal{E}_{IMU}$ are sets of edges on which we enforce limb length and orientation constraints, respectively. The unary potential $\phi_i^{\text{conf}}({J}_i, \mathcal{F})$ is computed based on $2$D pose heatmaps $\mathcal{F}$. The pairwise potential $\psi^{\text{limb}}({J}_m, {J}_n)$ and $\psi^{\text{IMU}}({J}_m, {J}_n)$ encode the limb length and orientation constraints. We describe each term in detail as follows.

\paragraph{Discrete State Space} We first estimate the $3$D location of the root joint by triangulation based on its $2$D locations detected in all views. Note that this step is usually very accurate because the root joint can be detected in most times. Then the state space of the $3$D pose is within a $3$D bounding volume centered at the root joint. The edge length of the volume is set to be $2000$mm which is large enough to cover every body joint. The volume is discretized by an $N \times N \times N$ regular grid $\mathcal{G}$. Each joint can take one of the bins of the grid as its $3$D location. Note that all body joints share the same state space $\mathcal{G}$ which consists of $N^3$ discrete locations (bins).

\paragraph{Unary Potential} Every body joint hypothesis, \ie, a bin in the grid $\mathcal{G}$, is defined by its $3$D position. We project it to the pixel coordinate system of all camera views using the camera parameters, and get the corresponding joint confidence/response from $\mathcal{F}$. We compute the average confidence/response over all camera views as the unary potential for the hypothesis.

\paragraph{Limb Length Potential}
For each pair of joints (${J}_m$,${J}_n$) in the edge set $\mathcal{E}_{limb}$, we compute the average distance $\Tilde{l_{m,n}}$ on the training set as limb length prior. During inference, the limb length pairwise potential is defined as:
\begin{equation}
    \psi^{\text{limb}}({J}_m, {J}_n) = 
    \left\{
                \begin{array}{ll}
                  1, \quad \text{if} \quad |l_{m,n} - \Tilde{l_{m,n}}| \leq\epsilon,\\
                  0, \quad \text{otherwise}
                \end{array}
              \right.,
\end{equation}
where $l_{m,n}$ is the distance between ${J}_m$ and ${J}_n$.
The pairwise term favors $3$D poses having reasonable limb lengths. In our experiments, $\epsilon$ is set to be $150$mm.

\paragraph{Limb Orientation Potential}
We compute the dot product between the limb orientations of the estimated pose and the IMU orientations as the limb orientation potential
\begin{equation}
    \psi^{\text{IMU}}({J}_m, {J}_n)= \frac{{J}_m-{J}_n}{\|{J}_m-{J}_n\|_2} \cdot {o}_{m,n},
\end{equation}
where $o_{m,n}$ is the orientation (represented as a directional vector) of the limb measured by the IMU. This term favors poses whose limb orientations are consistent with the IMUs. We also experimented with the hard orientation constraint similar to what we did for limb length, but this soft limb orientation constraint gets better performance. A $3$D pose estimator without/with \emph{orientation potential} will be termed as \textbf{\emph{PSM}} and \textbf{\emph{ORPSM}}, respectively.

\begin{figure}[ht]
\centering
\includegraphics[width=0.45\textwidth]{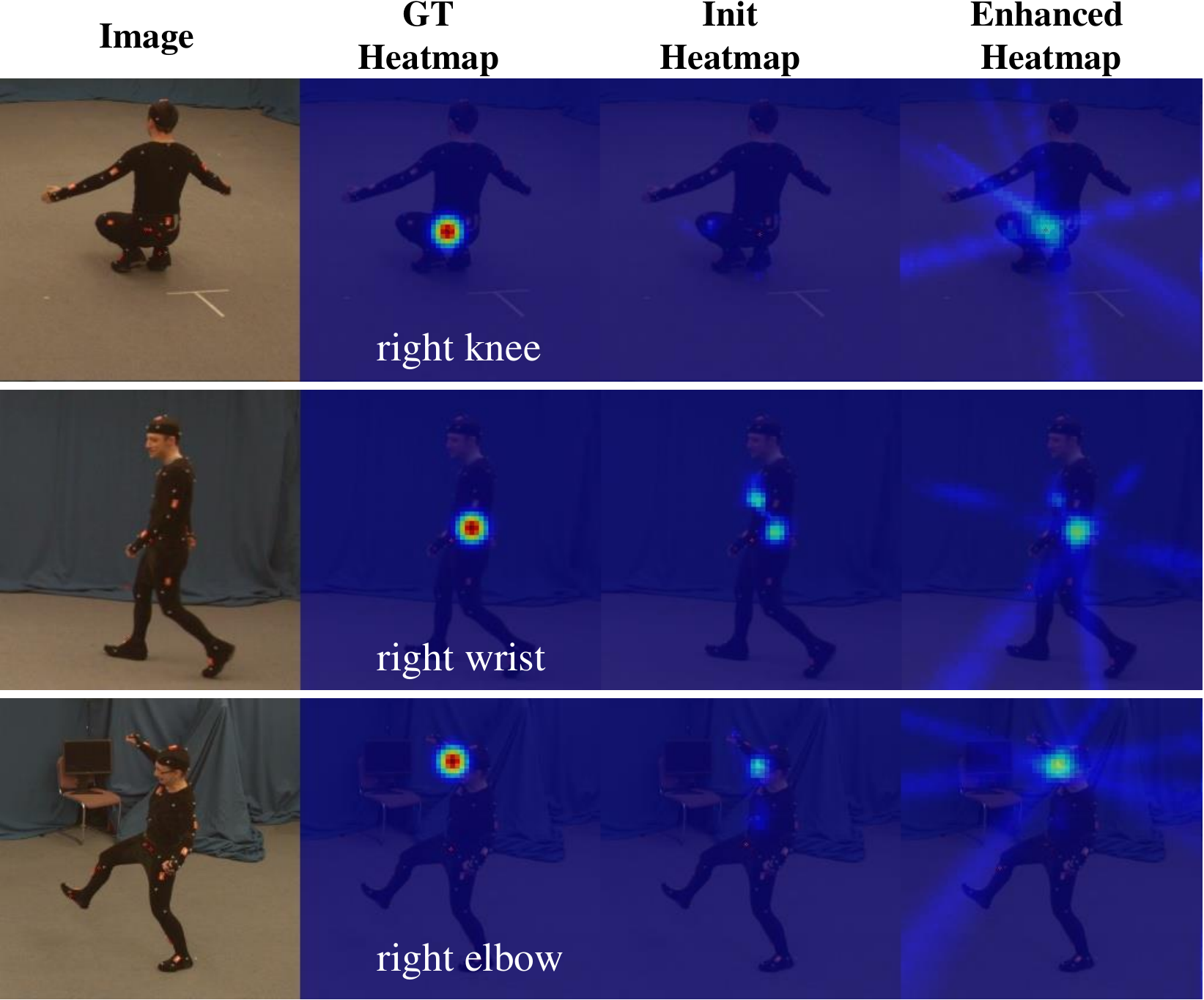}
\caption{Three sample heatmaps estimated by \emph{ORN}. The initially estimated heatmaps without fusion are inaccurate. After fusing multi-view features, the ``Enhanced Heatmap'' localizes the correct joints. Note there are blurred lines in the ``Enhanced Heatmap'' with each corresponding to the confidence contributed from one camera view. The lines intersect at the correct location.}
\label{fig:visualization_hm}
\end{figure}

\begin{figure*}[htbp]
\centering
\includegraphics[width=0.85\textwidth]{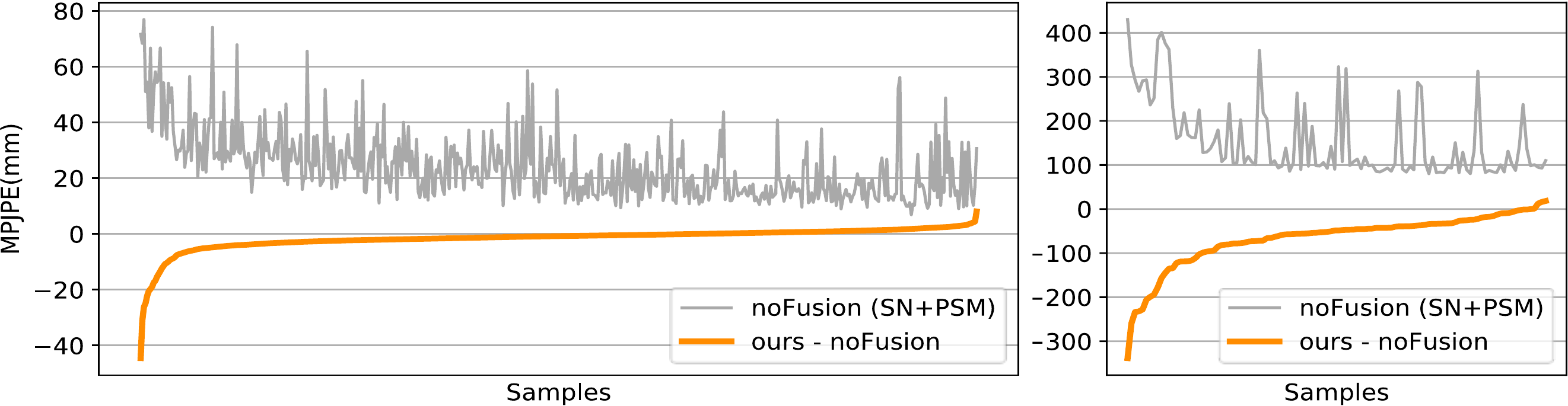}
\caption{The grey line shows the $3$D MPJPE error of the \emph{noFusion} approach. The orange line shows the error difference between \emph{our method (ORN+ORPSM)} and \emph{noFusion}. If the orange line is below zero, it means \emph{our method} has smaller errors. We split the testing samples into two groups according to the error scale of \emph{noFusion}. The first group includes the samples whose errors are smaller than $80$mm (shown in the left figure). The second group includes the rest of the samples (shown in the right figure). The samples are sorted by the orange line for the sake of readability.}
\label{fig:ifusion-vs-nofusion}
\end{figure*}

\paragraph{Inference}
We maximize the posterior probability, \ie Eq. (\ref{eq:psm}), over the discrete state space by the dynamic programming algorithm. In general, the complexity grows quadratically. In order to improve the speed, we adopt a recursive variant of PSM \cite{qiu2019cross} which iteratively refines the $3$D poses. In practice, it takes about $0.15$ seconds to estimate one $3$D pose on a single Titan Xp GPU.

\section{Datasets and Metrics}
\paragraph{Total Capture \cite{trumble2017total}} To the best of our knowledge, this is the only benchmark providing images, IMUs and ground truth $3$D poses. It places $8$ cameras in the capture room to record the human motion. We use four of them ($1$, $3$, $5$ and $7$) in our experiments for efficiency reasons. The performers wear $13$ IMUs. We use eight of them as shown in Figure \ref{fig:geometry} (c). There are five subjects performing four actions including Roaming(\textbf{R}), Walking(\textbf{W}), Acting(\textbf{A}) and Freestyle(\textbf{FS}) with each repeating $3$ times. Following the previous work \cite{trumble2017total}, we use Roaming 1,2,3, Walking 1,3, Freestyle 1,2 and Acting 1,2 of \emph{Subjects 1,2,3} for training our $2$D pose estimator. We test on Walking 2, Freestyle 3 and Acting 3 of all subjects.

\paragraph{H36M \cite{ionescu2014human3}}
To validate the general applicability of our approach, we also conduct experiments on the H36M dataset. Since this dataset does not provide IMUs, we create virtual IMUs (limb orientations) using the ground truth $3$D poses for both training and testing, and only show proof-of-concept results. Following the dataset conventions, we use subjects $1,5,6,7,8$ for training and subjects $9,11$ for testing. We train a single model for all actions.

\paragraph{Metrics} 
The Percentage of Correct Keypoints (PCK) metric is used for $2$D pose evaluation. Specifically, PCKh@$t$ measures the percentage of the estimated joints whose distance from the ground-truth joints is smaller than $t$ times of the head length. Previous works report results when $t$ is $\frac{1}{2}$. In our experiments, we provide results when $t$ is set to be $\frac{1}{2}$, $\frac{1}{6}$ and $\frac{1}{12}$, respectively, in order to understand our approach more comprehensively. We use the Mean Per Joint Position Error (MPJPE) for $3$D pose evaluation \cite{ionescu2014human3}. It computes the distance between estimated poses and the ground truth poses. We report the average error over all joints and all instances.

\section{Experimental Results}
\begin{figure*}[]
\centering
\includegraphics[width=0.93\textwidth]{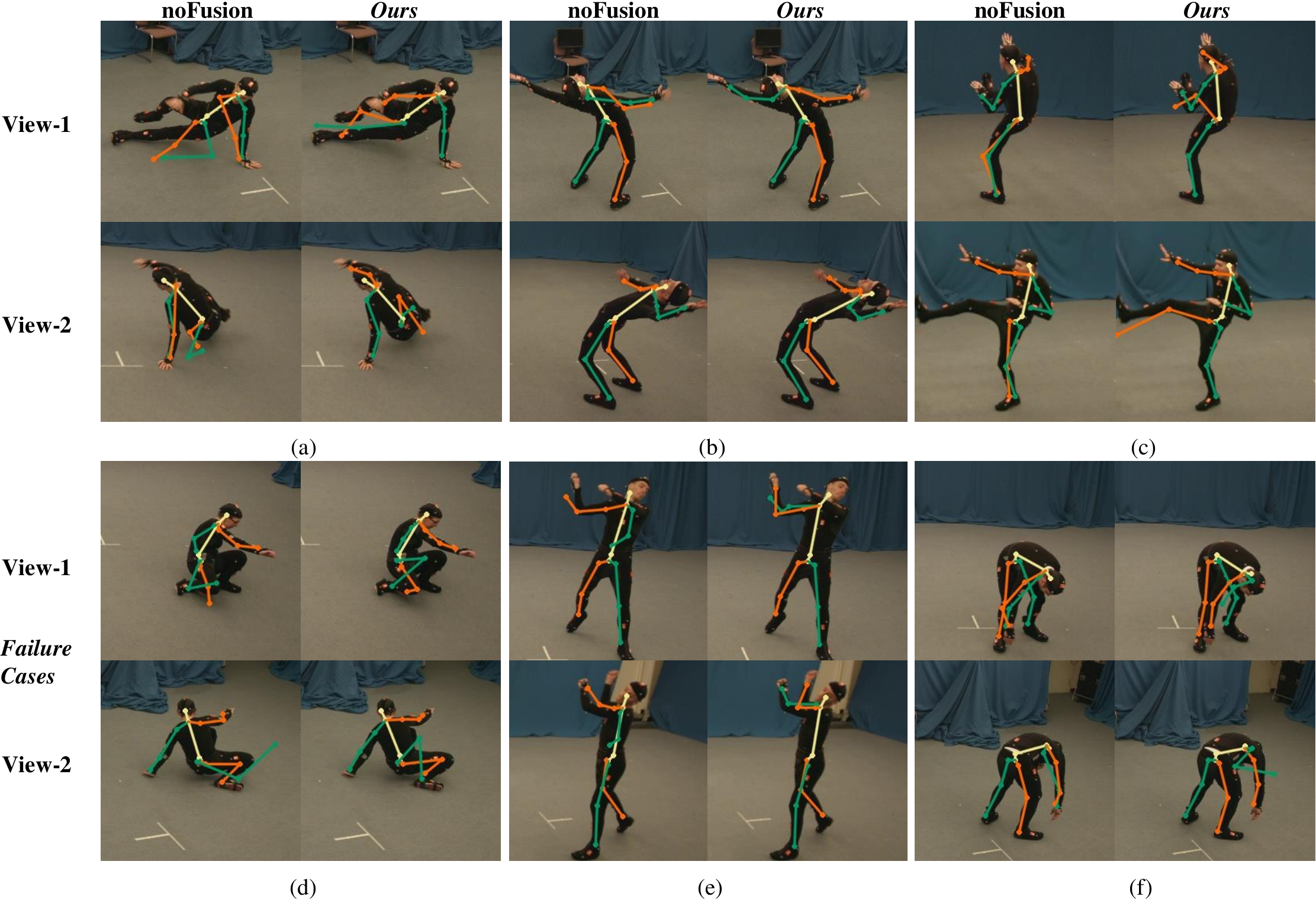}
\caption{Sample $3$D poses estimated by our approach and \emph{noFusion}. We project the estimated $3$D poses to the images and draw the skeletons.
Left and right limbs are drawn in green and orange colors, respectively. (a-c) show examples when our method improves over \emph{noFusion}. 
(d-f) show three failure cases.
These rare cases mainly happen when both joints of a limb have large errors.}
\label{fig:visual}
\end{figure*}

\subsection{$2$D Pose Estimation}
Table \ref{table:detectionrate} shows the $2$D pose estimation results of \emph{ORN} and the baseline \emph{SimpleNet (SN)}. We keep \emph{SN} the same as \emph{ORN} except it does not perform fusion. We can see from the table that when the threshold $t$ is set to be $\frac{1}{2}$ as in most previous works, \emph{ORN} outperforms \emph{SN} by a large margin. The improvement for wrist, elbow and knee joints is most significant because they are frequently occluded by human body in the dataset. Figure \ref{fig:visualization_hm} shows some examples explaining how our approach improves localization  over the baseline. For example, in the first example, initially, the right knee joint is not correctly detected because it is occluded by human body. Fusing the features from hip in multiple camera views helps localize it correctly. 

We notice that the improvement (brought by visual-inertial fusion) on the hip joints is small. There are two possible reasons for the phenomenon. First, the hip joints are visible for most images in the dataset. So IMUs provide barely no additional information. Second, the hip joint detection rate for the baseline method is already very high so it leaves little room for improvement. The estimation results for the joints without IMUs, which are represented as \emph{``Others''} in the table, are similar for the baseline and our approach, which is expected.

\begin{table*}[]
\center
\caption{$3$D pose estimation errors ($mm$) of different variants of our approach on the Total Capture dataset. ``Mean (six)'' is the average error over the six joint types. ``Others'' is the average error over the rest of the joints. ``Mean (All)'' is the average error over all joints.}
\label{table:ablation}
\begin{tabular}{cc||ccccccc||c||c}
\toprule
2D & 3D  & Hip  & Knee & Ankle & Shoulder & Elbow & Wrist & \emph{Mean (Six)} & Others & Mean (All) \\ \hline
SN      & PSM             & \textbf{17.2} & 35.7 & 41.2 & 50.5 & 54.8 & 56.8 & 37.1 & 20.3 & 28.3\\
\textbf{\emph{ORN}}  & PSM             & 17.4 & 29.9 & 35.2 & 49.6 & 44.2 & 45.1 & 32.8 & 20.4 & 25.4 \\
SN      & \textbf{\emph{ORPSM}}        & 18.3 & 25.8 & 34.0 & 44.8 & 44.2 & 49.8 & 32.1 & 19.9 & 25.5\\
\textbf{\emph{ORN}}  & \textbf{\emph{ORPSM}}        & 18.5 & \textbf{24.2} & \textbf{30.1} & \textbf{44.8} & \textbf{40.7} & \textbf{43.4} & \textbf{30.2} & {19.8} & {24.6} \\
\toprule
\end{tabular}
\end{table*}

\begin{table*}[]
\center
\caption{$3$D pose estimation errors MPJPE ($mm$) of different methods on the Total Capture dataset. ``Aligned'' means whether we align the estimated $3$D poses to the ground truth poses by Procrustes. }
\label{table:totalcapture}
\begin{tabular}{l c c c  c c c c c c c}
\toprule
Approach  & IMUs & Temporal & Aligned & \multicolumn{3}{c}{Subjects(S1,2,3)} & \multicolumn{3}{c}{Subjects(S4,5)} &  Mean \\ 
& & & & W2 & A3  & FS3 & W2 & A3 & FS3 & \\ \hline
PVH \cite{trumble2017total}  & & & & 48.3 & 94.3 & 122.3 & 84.3 & 154.5 & 168.5 & 107.3 \\
Malleson \etal \cite{malleson2017real} & \checkmark & \checkmark & & - & - & 65.3 & - & 64.0 & 67.0 & - \\
VIP \cite{von2018recovering} & \checkmark & \checkmark & \checkmark & - & - & - & - & - & - & 26.0 \\
LSTM-AE \cite{trumble2018deep} & & \checkmark & & \textbf{13.0} & 23.0 & 47.0 & \textbf{21.8} & 40.9 & 68.5 & 34.1 \\
IMUPVH \cite{gilbert2019fusing} & \checkmark & \checkmark & & 19.2 & 42.3 & 48.8 & 24.7 & 58.8 &  61.8 & 42.6\\ 
Qiu \etal \cite{qiu2019cross} & &  & & 19.0 & 21.0 & 28.0 & 32.0 & 33.0 &54.0 & 29.0 \\
\hline
\emph{SN + PSM} & & & & 14.3 & 18.7 & 31.5 & 25.5 & 30.5 & 64.5 & 28.3 \\
\emph{SN + PSM} & & & \checkmark & 12.7 & 16.5 & 28.9 & 21.7 & 26.0 & 59.5 & 25.3 \\\hline
\emph{ORN + ORPSM} & \checkmark & & & 14.3 & \textbf{17.5} & \textbf{25.9} & {23.9} & \textbf{27.8} & \textbf{49.3} &\textbf{24.6} \\ 

\emph{ORN + ORPSM} & \checkmark & & \checkmark & 12.4 & 14.6 & 22.0 & 19.6 & 22.4 & 41.6 & 20.6 \\
\toprule
\end{tabular}
\end{table*}

When we use a more rigorous threshold, for example when $t=\frac{1}{12}$, the detection rate for hip drops from $87.6\%$ to $85.3\%$ (SN vs. ORN). There are two reasons for this phenomenon: (1) the detection rate for hip is already very high for \emph{SN}, leaving little space for improvement; (2) IMUs often have small noises which may affect fusion precision. This conclusion is supported by the subsequent experimental results on the H36M dataset: when we use GT IMUs, the detection rate also improves for hip. Actually, even on the Total Capture, the impact also becomes small when we use a larger threshold. For example, when the threshold is set to be $\frac{1}{6}$, the accuracy of \emph{ORN} is slightly better than \emph{SN} ($97.7\%$ vs. $97.5\%$).

\begin{table*}[!htbp]
\center
\caption{$3$D pose estimation error ($mm$) on the H36M dataset. We use virtual IMUs in this experiment.
We show results for the six joints which are affected by IMUs. ``Mean (six)'' is the average error over the six joint types. ``Others'' is the average error over the rest of the joints. ``Mean (All)'' is the average error over all joints.}
\label{table:h36m_mpjpe}
\begin{tabular}{l||ccccccc||c||c}
\toprule
   Methods & Hip  & Knee & Ankle & Shoulder & Elbow & Wrist & \emph{Mean (Six)} & Others & Mean (All)\\ \hline
\emph{noFusion (SN + PSM)} & 23.2	&28.7&	49.4&	29.1&	28.4&	32.3&	31.9&	18.3 & 27.9 \\
\emph{ours (ORN + ORPSM)} & \textbf{20.6}	& \textbf{18.6} &	\textbf{28.2} &	\textbf{25.1} &	\textbf{21.8} &	\textbf{24.2} &	\textbf{23.1} &	18.3 & {21.7} \\
\toprule
\end{tabular}
\end{table*}

We also evaluate the impact of cross-view fusion in \emph{ORN}. As can be seen in Table \ref{table:detectionrate}, the multi-view fusion outperforms the same-view fusion consistently which validates its effectiveness. In addition, we find that the improvement is larger when we use a more rigorous threshold $t$. The results suggest that multi-view feature fusion helps localize the joints more precisely.

\subsection{$3$D Pose Estimation}
We first evaluate our $3$D pose estimator through a number of ablation studies. Then we compare our approach to the state-of-the-arts. Finally, we present results on the H36M dataset validating the generalization capability of the proposed approach.

\paragraph{Ablation Study}
We denote the baseline which uses \emph{SN} and \emph{PSM} to estimate $2$D and $3$D pose as \textbf{\emph{noFusion}} baseline. The main results are shown in Table \ref{table:ablation}. 
First, using \emph{ORN} consistently decreases the $3$D error no matter what $3$D pose estimators we use. In particular, the improvement on the elbow and wrist joint is as large as $10$mm when we use PSM as the $3$D estimator. This significant error reduction is attributed to the improved $2$D poses. 
Figure \ref{fig:visual} (a-c) visualize three typical examples where \emph{ORN} gets better results: we project the estimated $3$D poses to the images and draw the skeletons. It is guaranteed that if the 2D locations are correct for more than one view, then the 3D joint location is at the correct position. We also plot the $3$D error of every testing sample in Figure \ref{fig:ifusion-vs-nofusion}. Our approach improves the accuracy for most cases because the orange line is mostly below zero. See the caption of the figure for the meanings of the lines. In addition, we can see that the improvement is larger when the \emph{noFusion} baseline has large errors. There are a small number of cases where fusion does not improve joint detection results as shown in Figure \ref{fig:visual} (d-f).

Second, from the second and third rows of Table \ref{table:ablation}, we can see that using ORPSM alone achieves a similar $3$D error as ORN alone. This means $3$D fusion is related to $2$D fusion in some way--- although $3$D fusion does not directly improve the $2$D heatmap quality, it uses $3$D priors to select better joint locations having both large responses as well as small discrepancy with respect to the prior structures. But in some cases, for example, when the responses at the correct locations are too small, using the $3$D prior is not sufficient. This is verified by the experimental results in the fourth row --- if we enforce $2$D and $3$D fusion simultaneously, the error further decreases to $30.2$mm. It suggests the two components are actually complementary.

\paragraph{State-of-the-arts } Finally we compare our approach to the state-of-the-arts on the Total Capture dataset. The results are shown in Table \ref{table:totalcapture}. First, we can see that \emph{IMUPVH} \cite{gilbert2019fusing} which uses IMUs even gets worse results than \emph{LSTM-AE} \cite{trumble2018deep} which does not use IMUs. The results suggest that getting better visual features is actually more effective than performing late fusion of the (possibly inaccurate) $3$D poses obtained from images and IMUs, respectively. Our approach, which uses IMUs to improve the visual features, also outperforms \cite{gilbert2019fusing} by a large margin.

The error of the state-of-the-art is about $29$mm \cite{qiu2019cross} which is larger than $24.6$mm of ours. This validates the effectiveness of our IMU-assisted early visual feature fusion. Note that the error of VIP \cite{von2018recovering} is obtained when the $3$D pose estimations are aligned to ground truth which should be compared to $20.6$mm of our approach.

We notice that the error of our approach is slightly larger than \cite{trumble2018deep} for the ``W2 (walking)'' action. We tend to think it is because LSTM can get significant benefits when it is applied to \emph{periodic} actions such as ``walking''. This is also observed independently in another work \cite{gilbert2019fusing}. Besides, the error $14.3$mm for ``W2'' of Subject 1,2,3 is not further reduced after fusion since \textit{noFusion} method has already achieved extraordinarily high accuracy.

\paragraph{Generalization} To validate the wide applicability of our approach, we conduct experiments on the H36M dataset \cite{ionescu2014human3}. The results of different methods are shown in Table \ref{table:h36m_mpjpe}. We can see that our approach (ORN+ORPSM) consistently outperforms the baseline \emph{noFusion} which validates its general applicability. In particular, the improvement is significant for the Ankle joint which is often occluded. Since we use the ground truth IMU orientations in this experiment, the results are not directly comparable to other works.

\section{Summary and Future Work}
We present an approach for fusing visual features through IMUs for $3$D pose estimation. The main difference from the previous efforts is that we use IMUs in a very early stage. We evaluate the approach through a number of ablation studies, and observe consistent improvement resulted from the fusion. As the readings from the IMUs usually have noises, our future work will focus on learning a reliability indicator, for example based on temporal filtering, for each sensor to guide the fusion process.

\small
\bibliographystyle{ieee_fullname}
\bibliography{egbib}

\end{document}